\documentclass[10pt,twocolumn,letterpaper]{article}

\usepackage{cvm}
\usepackage{times}
\usepackage{epsfig}
\usepackage{graphicx}
\usepackage{amsmath}
\usepackage{amssymb}
\usepackage[utf8]{inputenc}
\usepackage{multirow} 
\usepackage{booktabs} 
\usepackage{multirow} 

\usepackage[pagebackref=true,breaklinks=true,letterpaper=true,colorlinks,bookmarks=false]{hyperref}

\newcommand{\keywords}[1]{{\bf \emph{Keywords: #1}}}
\cvmfinalcopy

\def\cvmPaperID{699} 

\ifcvmfinal\fi
\begin{document}

\title{No Training, Better Flights: Test-Time Scaled VLMs for UAV Navigation}

\author{First Author\\
Institution1\\
Institution1 address\\
{\tt\small firstauthor@i1.org}
\and
Second Author\\
Institution2\\
First line of institution2 address\\
{\small\url{http://www.author.org/~second}}
}

\author{
Feinan Cheng\\
Shandong University\\
Weihai, China\\
{\tt\small 202437608@mail.sdu.edu.cn}
\and
Dongliang Xu\\
Shandong University\\
Weihai, China\\
{\tt\small xudongliang@sdu.edu.cn}
\and
Wenli Nong\\
Shandong University\\
Weihai, China\\
{\tt\small 202200630135@mail.sdu.edu.cn}
\and
Zhiheng Zhang\\
Shandong University\\
Weihai, China\\
{\tt\small 202537837@mail.sdu.edu.cn} 
\and
Ang Liu\\
China University of Petroleum (East China)\\
Qingdao, China\\
{\tt\small z24070078@s.upc.edu.cn}
\and
Tianyu Wang\\
MBZUAI\\
Abu Dhabi, United Arab Emirates\\
{\tt\small tianyu.wang@mbzuai.ac.ae}
\and
Yue Yao\\
Shandong University\\
Weihai, China\\
{\tt\small yue.yao@anu.edu.au}
}

\maketitle

\begin{abstract}    
     Test-time scaling offers a promising method to improve the inference performance of Vision-Language Models (VLMs) without additional training. Existing approaches to vision-language navigation (VLN) for Unmanned Aerial Vehicle (UAV) typically relies on a single inference pass, which can falter in complex environments by producing suboptimal or unsafe trajectories. In this paper, we explore a simple and effective approach to apply test-time scaling to VLN for UAV. We enhance navigation reasoning through an iterative refinement process that requires no extra model training, guiding the model to re-evaluate its initial navigation plan for better accuracy and safety. Our method first prompts the model to generate multiple parallel candidates and then performs a self-correction step, achieving deeper and more robust planning without changing the underlying model. To further strengthen decision-making, we design a multi-criteria scoring function to evaluate the refined candidates based on safety, goal alignment, and forward-progress. This simple yet powerful combination enables a frozen UAV navigation VLMs to self-correct and generate more accurate and reliable flight plans, achieving SOTA performance in this task.
\end{abstract}


\section{Introduction}


\begin{figure*}[t] 
    \centering
    \includegraphics[width=1\textwidth]{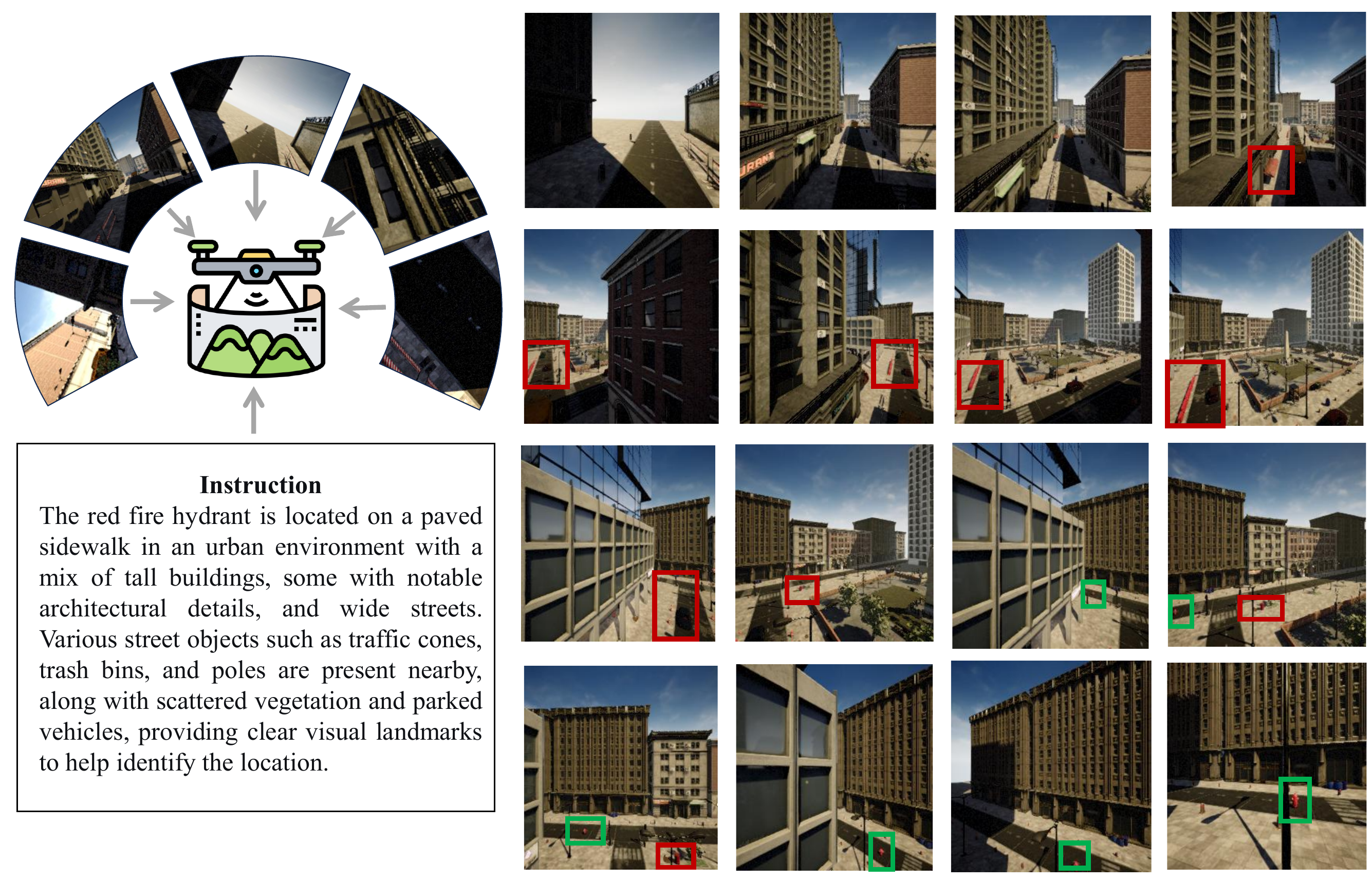} 
    \caption{
    \textbf{Task definition for UAV navigation.} This set of images shows the target recognition process for a UAV in a simulation environment. The red boxes represent negative samples, which are distracting objects whose color or shape might cause confusion (such as traffic cones and trash cans). The green boxes represent positive samples, which are fire hydrants that need to be located.
    }
    \label{fig:figure3}
    \vspace{0.5em}
\end{figure*}
    
    Recently, the application of VLMs to the autonomous navigation of UAVs has emerged as a research direction of widespread interest. The core objective of this research is to establish a more intuitive and flexible paradigm for human-robot interaction. This would enable non-expert users to deploy and control UAVs for complex tasks using simple natural language instructions, thereby significantly lowering the operational threshold. This technology presents a wealth of application scenarios, including automated logistics, large-scale infrastructure inspection, precision agriculture, and post-disaster search and rescue. It is poised to significantly enhance both operational efficiency and safety in these critical domains. Figure~\ref{fig:figure3} provides an example that visually illustrates the complexity of such a task. The UAV needs to interpret human-generated instructions and accurately identify a specific red fire hydrant in a dense urban environment filled with visual distractors such as red trash cans and traffic cones. Consequently, developing intelligent navigation agents capable of accurately understanding high-level human commands and making robust and deliberate decisions has become a recognized and crucial challenge at the intersection of Artificial Intelligence and robotics.
    
    Foundational work in this area has successfully demonstrated the feasibility of VLMs-based navigation agents, providing the research community with essential platforms and baseline models. For instance, Wang \emph{et al}.~\cite{Wang2024} pioneered the development of the OpenUAV platform, which offers a realistic simulation environment for UAV-based VLN, and established a baseline VLMs-based navigation agent. Concurrently, other studies, such as the work by Fan \emph{et al}.~\cite{Fan2022}, have explored aerial vision-and-dialog navigation. These pioneering efforts have successfully validated the fundamental feasibility of VLMs in translating language instructions into flight trajectories. However, these studies predominantly employ a single-step decision-making model, where the agent generates an action directly from the current multimodal input. While this approach is computationally efficient, its decision-making process resembles an instantaneous response. Consequently, its reliability in complex and dynamic environments remains an area requiring significant improvement.
    
    Although existing methods have validated its feasibility, their core challenge lies in the limitation of their planning capabilities. These navigation agents lack the capacity for more deliberate planning prior to action, such as evaluating multiple possibilities and performing self-correction. This deficiency leads to insufficient reliability when facing complex or ambiguous instructions, often resulting in the generation of suboptimal paths. To enhance the decision-making quality and robustness of UAV navigation agents, we introduce the \textit{test-time scaling} paradigm, which has proven effective in the broader domain of large language models. This paradigm boosts a model's reasoning depth and breadth by allocating more computational resources at the inference stage. It is primarily divided into two main strategies: (1) \textbf{sequential scaling}, which utilizes techniques like the Chain-of-Thought method developed by Wei \emph{et al}.~\cite{Wei2022} and the Self-Refine framework by Madaan \emph{et al}.~\cite{Madaan2023} to optimize a single decision chain; and (2) \textbf{parallel scaling}, which employs methods like the Tree-of-Thoughts from Yao \emph{et al}.~\cite{Yao2023} to explore multiple reasoning paths concurrently. We contend that single-dimensional scaling has its own limitations: purely sequential scaling, while deepening the reasoning process, may become trapped in a local optimum; purely parallel scaling, despite its broad exploration, offers limited depth for each individual path.

    Therefore, this research aims to explore an inference framework that integrates parallel exploration with serial refinement to improve the quality of planning and decision-making for VLMs in UAV navigation tasks. To this end, we have designed a ``breadth-first, depth-optimized'' inference strategy. First, the model generates multiple parallel candidate trajectories through parallel exploration, which prevents premature convergence on a single, suboptimal solution. Subsequently, each candidate trajectory undergoes a serial, iterative self-correction phase, where the model is prompted to critically re-evaluate its initial plan. To select the optimal solution from these refined candidates, we introduce a multi-criteria scoring function that holistically assesses each candidate path based on its safety, goal alignment, and forward progress. Our experiments validate the independent contributions of both the sequential and parallel components. Ultimately, we demonstrate that our proposed method achieves better navigation performance compared to single-dimensional scaling strategies and traditional baselines, while also verifying the positive correlation between navigation SR and the inference-time compute budget (token consumption).
    
    
    \section{Related Work}
    
    \textbf{UAV Navigation.} In recent years, research on drone navigation has focused on addressing two core challenges: maintaining precise positioning in environments with limited or denied GNSS signals, and ensuring flight safety in complex scenarios with dynamic obstacles \cite{chang2023review}. To address these challenges, early fundamental research has primarily explored multi-sensor fusion technology. One prominent technical direction is to suppress the accumulated drift of Visual-Inertial Odometry (VIO) through sensor fusion \cite{pritzl2023fusion}. Combining VIO with LiDAR or Ultra-Wideband (UWB) has been shown to be an effective means of achieving robust relative positioning. Further research has shown that a comprehensive positioning architecture integrating VIO  \cite{bednavr2022deployment}, LiDAR odometry, and UWB anchors can provide highly robust navigation performance in extremely challenging environments \cite{diez2024time}.
    
    However, traditional sensor-level fusion approaches struggle to address long-range navigation tasks requiring advanced semantic understanding. To address this shortcoming, cutting-edge research is integrating the cognitive capabilities of large language models (LLMs) into the navigation decision loop. This shift from pure perception to cognitive decision-making aims to enhance the planning and reasoning capabilities of intelligent agents. For example, studies have combined LLMs with vision-based obstacle prediction, demonstrating their effectiveness in improving the navigation safety of drones in crowded environments \cite{zhong2024safer}.
    In the realm of high-speed flight, while systems combining onboard sensing with deep reinforcement learning (RL) have demonstrated championship-level performance in drone racing \cite{kaufmann2023champion,hanover2024autonomous}, end-to-end reinforcement learning methods are still primarily used to explore basic point-to-point operations \cite{oyinlola2025reinforcement,guo2025autonomous}. On the other hand, advanced environmental representation techniques are providing a new foundation for intelligent navigation. Researchers are applying advanced 3D reconstruction techniques such as NeRF and 3D Gaussian Splatting (3DGS) to aerial viewpoints to support visual localization and trajectory optimization with collision avoidance capabilities \cite{jia2024drone}.
    Research in multi-agent collaboration and advanced human-computer interaction continues to deepen. For example, some work has leveraged UWB ranging to achieve collaborative relative positioning of Micro-Aircraft (MAV) swarms \cite{liu2024cooperative}, or proposed ``drone-guiding-drone'' navigation schemes \cite{pritzl2024drones}. Furthermore, frameworks such as AeroDuo \cite{wu2025aeroduo} further advance this research by separating environmental reasoning from fine-grained navigation using heterogeneous drones operating at different altitudes.
    At the same time, VLN is becoming a new research hotspot. The groundbreaking work AerialVLN \cite{liu2023aerialvln} facilitated the integration of language, perception, and control by providing simulators and datasets. Building on this foundation, new platforms and datasets such as UAV-Need-Help \cite{Wang2024} further advance this field by emphasizing realistic assisted target search and command tracking trajectories.


\begin{figure*}[t] 
    \centering
    \includegraphics[width=1\textwidth]{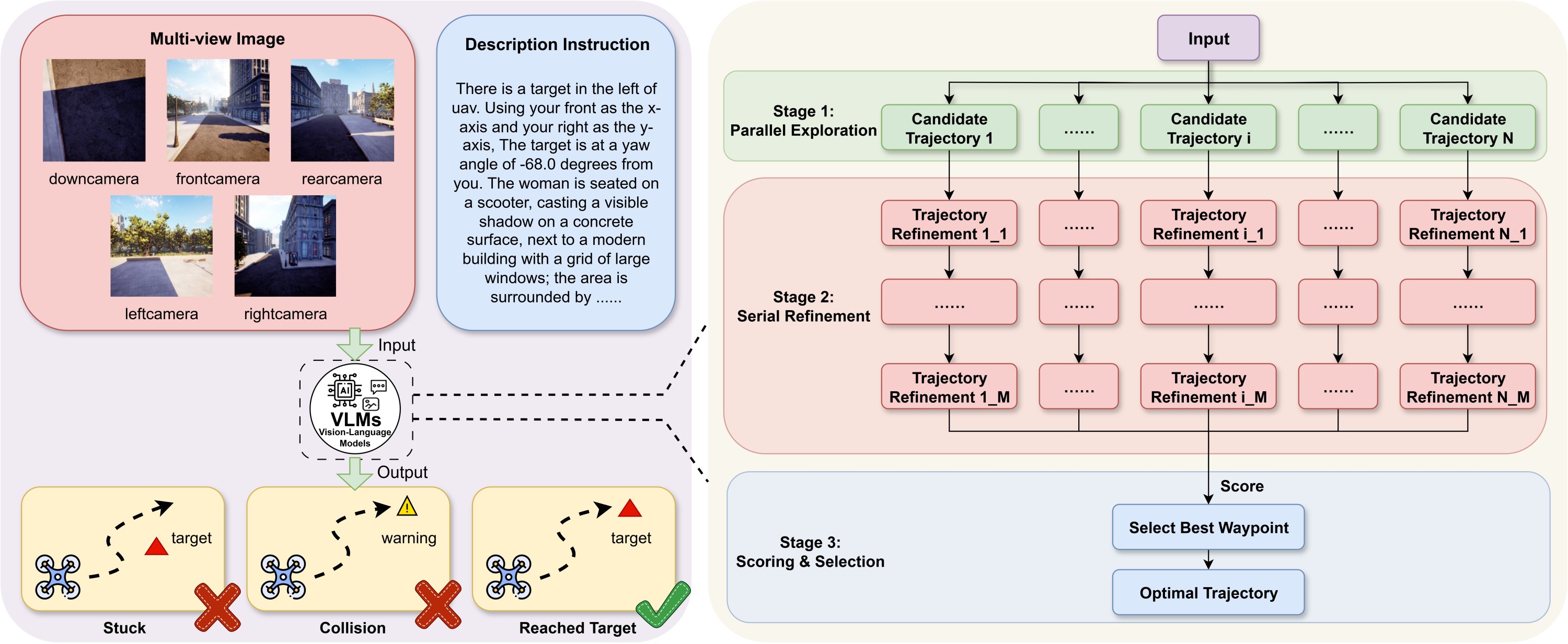} 
    \caption{
    \textbf{The Multi-Step Pipeline for Trajectory Generation and Refinement.} (Left) The VLMs processes multimodal inputs, including multi-view images and a language instruction, to produce a navigation output. This panel illustrates the potential outcomes of different planning qualities, ranging from mission failure (e.g., getting stuck or collision) to success. (Right) The core of the method is a three-stage processing pipeline. In Stage 1 (Parallel Exploration), the VLMs generates N distinct Candidate Trajectories based on the initial input. In Stage 2 (Serial Refinement), each candidate undergoes M rounds of iterative self-correction, where the VLMs re-evaluates and improves its own initial plan, resulting in a set of Refined Trajectories. Finally, in Stage 3 (Scoring \& Selection), all refined trajectories are evaluated by a scoring function, which selects the Optimal Trajectory to be executed by the UAV.
    }
    \label{fig:motivation}
\end{figure*}

\textbf{VLMs.} 
The field of VLMs is experiencing rapid development, with both model capabilities and efficiency continuously improving. This progress has led to the emergence of a dominant technical paradigm: efficiently bridging powerful pre-trained visual encoders with instruction-fine-tuned LLMs through a lightweight projection layer. This robust architectural foundation has further advanced research, extending the capabilities of VLMs from static image understanding to more complex dynamic multimodal reasoning tasks.
The evolution of such models was initially driven by work such as LLaVA, which established the effectiveness of fine-tuning visual instructions for single-image tasks such as caption generation and VQA \cite{liu2023visual}. Subsequently, the capabilities of VLMs rapidly expanded beyond static inputs. The development of the LLaVA family exemplifies this trend: from LLaVA-OneVision \cite{li2024llava}, which supports multiple images and videos, to LLaVA-NeXT-Interleave \cite{li2024llava0}, which can handle interleaved and multi-view inputs, the models' ability to handle dynamic and complex visual information has been systematically enhanced.
At the same time, contributions from the open source community are driving progress in this field along two key dimensions. On the one hand, open source suites such as InternVL-1.5 are significantly narrowing the gap with leading proprietary models by achieving superior performance in tasks such as document reasoning \cite{chen2024far}. On the other hand, model families such as Qwen2-VL focus on architectural innovations, introducing novel mechanisms such as dynamic resolution tokenization to improve computational efficiency and long-context understanding capabilities \cite{wang2024qwen2,bai2025qwen2}. 
The practical deployment of VLMs, especially in resource-constrained scenarios such as drones, faces severe computational efficiency challenges. Therefore, improving inference efficiency has become a key research topic. In this area, in addition to techniques such as dynamic resolution, a variety of technologies have emerged to optimize token processing, such as adaptive token pruning and compression (e.g., ATP-LLaVA \cite{ye2025atp} and TokenPacker \cite{li2025tokenpacker}). These methods significantly reduce computational overhead while minimizing accuracy loss, laying the foundation for model deployment on edge devices.
In summary, the collective advances in general VLMs in long-context processing, cross-modal understanding, and computational efficiency have created a mature technical foundation for applying these models to complex perception tasks requiring continuous decision-making \cite{song2025towards}, such as embodied intelligence. 

The value and potential of these advances are particularly evident in the cutting-edge field of VLN for aerial platforms, which also places higher demands on the models' deliberative reasoning capabilities.
Significant progress has been made in task design and evaluation methods for aerial VLN. The evolution of this research direction is clearly evident: from early foundational benchmarks such as AerialVLN \cite{liu2023aerialvln} to more recent, end-to-end language-guided flight systems that are closer to real-world applications, such as VLFly \cite{zhang2025grounded}. As research deepens, new research (such as FG-AVDN) has begun to focus on finer-grained grounding mechanisms to overcome the performance bottlenecks caused by the inaccurate alignment of commands and visual information in early datasets. By introducing entity-landmark alignment, it significantly improves the grounding accuracy of navigation commands \cite{su2025learning}.
In addition, recent research has begun to explore how to enhance the long-term decision-making capabilities of models. For example, FlightGPT uses the chain-of-thought reasoning of VLMs to achieve better planning \cite{cai2025flightgpt}. These diverse studies point to a clear trend: building a more scalable and robust visual-language navigation framework. However, existing work mainly achieves this goal by improving model architecture or optimizing training strategies. In contrast, our research has opened up a complementary research path: we focus on the inference stage and explore how to improve the inherent navigation performance of pre-trained models by extending the technology at test time without changing any pre-trained model parameters.


\textbf{Test time scaling}, which dynamically adjusts computational resources during inference, uses the ``budget forcing'' technique to control the computational load during the model inference process, thereby significantly improving the performance of the model in complex inference tasks without increasing training costs \cite{muennighoff2025s1}. This strategy has been explored across multiple application areas. For instance, lines of work on neural architecture search \cite{song2023efficient} and dynamic inference \cite{montello2025survey} support leveraging adaptive compute to improve performance on difficult inputs. 
This concept has also been validated in related fields. For example, a recent review of test-time adaptation systematically identifies a series of methods that share a core philosophy with test time scaling: they all aim to improve inference performance without retraining the model \cite{xiao2024beyond}. These diverse research directions all point to a key underlying principle: increasing computational input during the inference phase effectively unlocks and enhances the model's inherent reasoning capabilities \cite{snell2024scaling}. Therefore, integrating technology into UAV VLN tasks is a potential path to improve navigation performance.

\section{Method}

\subsection{Overall Framework: Test Time Explore--Refine--Select Navigation}

We propose an inference framework that applies the test time scaling paradigm to UAV navigation, designed as a three-stage \emph{Explore--Refine--Select} process. Our framework aims to significantly enhance the robustness and accuracy of VLMs when performing tasks in complex, dynamic environments. This core mechanism reduces the risk of making incorrect decisions based on single-gut thinking by introducing a systematic self-review and selection process.

As illustrated in Figure~\ref{fig:motivation}, our method consists of three tightly integrated stages: \textbf{Parallel Exploration}, \textbf{Serial Refinement}, and \textbf{Scoring \& Selection}. In the first stage, the model diverges and explores multiple candidate navigation points. In the second stage, it critically and iteratively reflects on and self-corrects each candidate. Finally, in the third stage, a comprehensive scoring function evaluates all optimized solutions to select the most reliable decision.
\subsection{Dissecting the Three-Stage Pipeline}

The following sections detail the design and implementation of the first stage, ``\textbf{Parallel Exploration}'', the second stage, ``\textbf{Serial Refinement}'', and the third stage, ``\textbf{Scoring \& Selection}''.


\begin{figure*}[t] 
    \centering
    \includegraphics[width=0.9\textwidth]{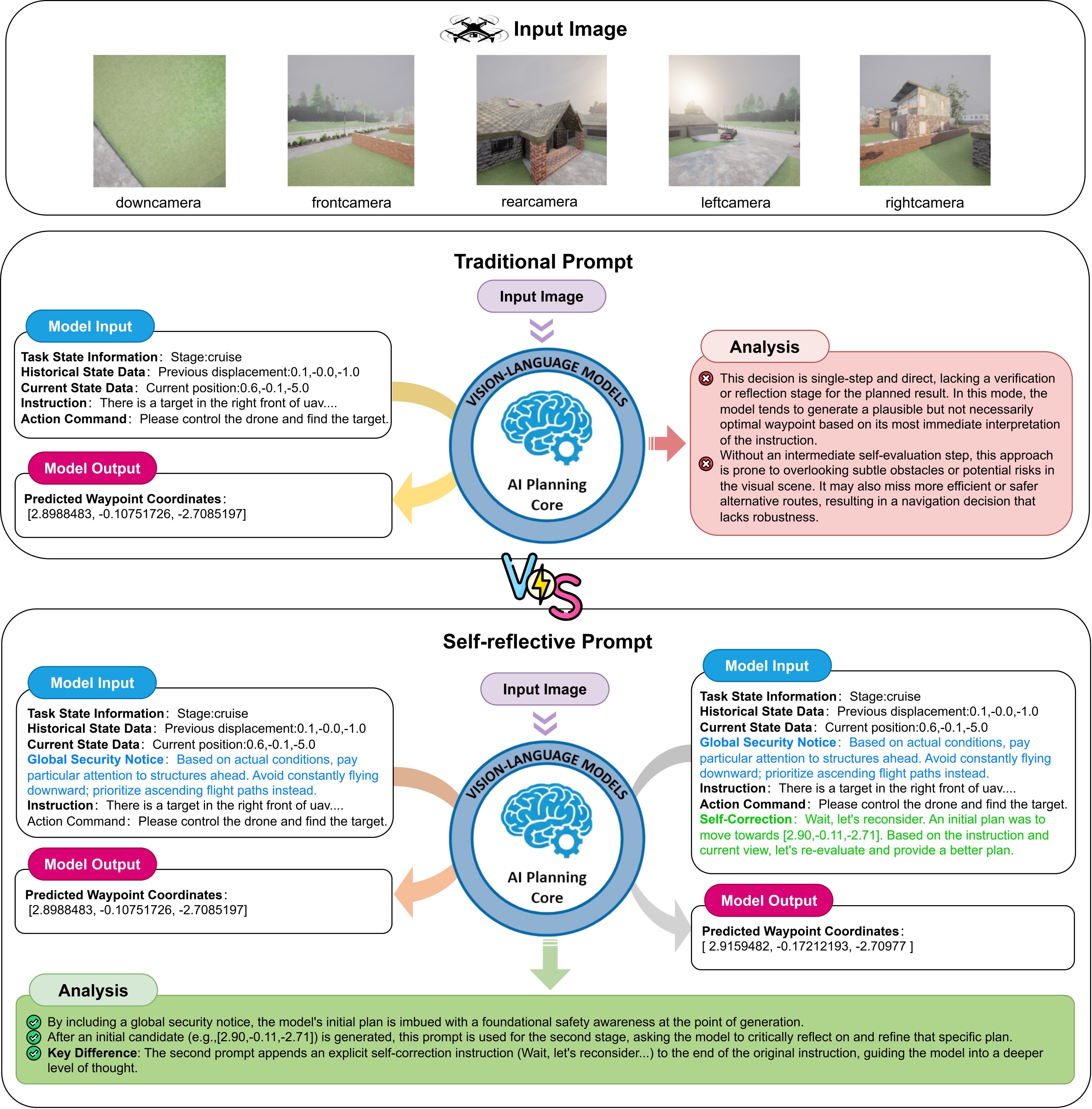} 
    \caption{
    \textbf{Comparison of the traditional prompt and self-reflective prompt workflows.} (Top) shows the single-step, direct generation process of the traditional Prompt. (Bottom) shows the``reflection-optimization'' process introduced by our self-reflective Prompt, where an initial plan is reevaluated to produce an optimized output.
    }
    \label{fig:example}
    \vspace{0.5em}
\end{figure*}


\subsubsection{Stage 1: Parallel Exploration}
\label{sec:parallel_exploration}
The process begins with a parallel exploration stage, designed to address the inherent uncertainties in complex navigation tasks. We formalize the initial multimodal input as a tuple $I = (V, L)$, where $V = \{v_1, v_2, \dots, v_k\}$ represents the visual observations from $k$ different camera views, and $L$ represents the user's natural language instruction.

In this phase, we are not seeking a single optimal solution. We model the VLMs as a function $F$ that takes a multimodal input $I$. By driving the VLMs to generate $N$ different candidate coordinates in parallel through $N$ independent inference calls to the model, we obtain $N$ initial candidate points:

\begin{equation}
    c_i = \mathcal{F}(I), \quad \forall i \in \{1, 2, \dots, N\}
    \label{eq:generation}
\end{equation}

Each initial candidate point $c_i$ is a coordinate point in three-dimensional space. The final output of this divergent thinking phase is a diverse set of initial candidates:
\begin{equation}
    C_{initial} = \{c_1, c_2, \dots, c_N\}
    \label{eq:initial_candidates}
\end{equation}
This set provides a broad foundation of initial hypotheses for the subsequent refinement and selection stages, thereby mitigating the risk of premature convergence on a suboptimal or unsafe path.


\subsubsection{Stage 2: Serial Refinement}
\label{sec:serial_refinement}
Following the exploration stage, each initial candidate point $c_i$ from the set $C_{initial}$ undergoes a serial refinement process, which forms the core of our iterative reasoning mechanism. The objective is to have the model critically re-evaluate and optimize its own preliminary plan in a `self-reflective` manner.

To achieve this, we introduce a self-reflective prompt strategy. As shown in Figure~\ref{fig:example}, in the traditional prompt word mode, the model generates predicted waypoint coordinates in a single step. Because this direct response model lacks a review process, the model can easily overlook subtle obstacles or potential risks in the visual scene, leading to potentially suboptimal decisions. In contrast, our self-reflective prompt model optimizes this decision-making process. First, through analysis of failure cases, we discovered that the UAV consistently flew downward during its initial phase and subsequently collided. Therefore, we added global safety instructions to the prompts. The model receives a preliminary prompt containing a global security notice, allowing it to more realistically generate its initial predicted waypoint coordinates, rather than simply flying downward. We then reinject these waypoint coordinates into the prompt along with an explicit self-correction instruction (``Wait, let's reconsider...''), prompting the model to critically reflect on its initial thinking. This reflection-refinement process guides the model into deeper reflection, ultimately producing more thoughtful, safer, and more reliable waypoint coordinates.

We model this process of reflection and correction as another function, $\mathcal{R}$. The input to this function comprises not only the original multimodal observations $I$ but also the initial point $c_i$ that is to be refined. In practice, we integrate $c_i$ into a specific reflective prompt template, $\mathcal{T}_{prompt}$, which linguistically guides the model to perform a re-evaluation. For instance:
\begin{quote}
    \textit{``Wait, let's reconsider. An initial plan was to move towards [{waypoints}]. Based on the instruction and current view, let's re-evaluate and provide a better plan.''}
\end{quote}
Here, \textit{[{waypoints}]} is the string representation of $c_i$. This self-correction step can be formalized as:
\begin{equation}
    c'_i = \mathcal{R}(I, c_i; \mathcal{T}_{prompt})
    \label{eq:refinement}
\end{equation}
where $c'_i$ is a more considered, refined point generated after M rounds of self-correction by the model. This process is executed for all $N$ candidate points in $C_{initial}$, ultimately producing a higher-quality set of refined candidates:
\begin{equation}
    C_{refined} = \{c'_1, c'_2, \dots, c'_N\}
    \label{eq:refined_candidates}
\end{equation}
The points in this set, having undergone the model's critical self-appraisal, typically show significant improvements in safety, feasibility, and goal-alignment compared to the initial candidates, providing a superior input for the final decision-making in Stage 3.


\subsubsection{Stage 3: Scoring \& Selection}
\label{sec:scoring_selection}

In the final stage of the pipeline, we introduce a convergent decision-making mechanism designed to adjudicate the optimal path from the refined set of candidate points, $C_{refined}$. To this end, we have designed a multi-criteria scoring function that holistically evaluates each candidate point.
The evaluation is primarily centered around three core dimensions: Safety, Goal-Alignment, and Forward-Progress.
For any given candidate point $c'_i \in C_{\text{refined}}$, its total score, $\text{Score}_{\text{total}}(c'_i)$, i is calculated using the following weighted sum:
\begin{equation}
    Score_{total}(c'_i) = w_{obs} \cdot S_{obs} + w_{tar} \cdot S_{tar} + w_{prog} \cdot S_{prog}
    \label{eq:total_score}
\end{equation}
where $S_{obs}, S_{tar}, \text{and } S_{prog}$ represent the Safety Score, Goal-Alignment Score, and Forward-Progress Score, respectively. The terms $w_{obs}, w_{tar}, \text{and } w_{prog}$ are their corresponding weighting coefficients, which we set to 0.5, 0.3, and 0.2 to ensure that safety is the primary consideration in UAV flight. We assign the largest weight to Safety to reduce collision risk. We use smaller weights for Goal Alignment to support task completion and for Forward Progress to avoid aggressive long-step proposals. The specific definitions for each scoring component are as follows.

\textbf{Safety Score ($S_{obs}$).} This score is designed to quantify the collision risk of a path, prioritizing UAV navigation safety. We assess the safety of the environment by analyzing depth image data from multiple directions carried by the UAV.Specifically, we select a set of $K$ key depth views (front, back, left, right, bottom), denoted as $\mathcal{D} = \{D_1, \dots, D_K\}$.
To mitigate image edge distortion and noise, we focus only on the central region of interest (RoI) of each depth map $D_k \in D$, denoted as $R_k$. We first find the minimum depth value within each $R_k$, which represents the nearest obstacle in that direction. The final safe distance $d_{safe}$ is defined as the minimum of the minimum depth values in all $R_k$:
\begin{equation}
    d_{safe} = \min_{k \in \{1,...,K\}} \left( \min_{p \in R_k} D_k(p) \right)
    \label{eq:safety_distance}
\end{equation}
where $p$ is the pixel point in $R_k$. In order to make the score drop sharply when $d_{safe}$ is small and increase slowly when it is large, we use a logarithmic function to perform a nonlinear transformation. The final safety score is:
\begin{equation}
    S_{obs} = \log(d_{safe} + 1)
    \label{eq:safety_score}
\end{equation}
This function ensures that when the UAV approaches an obstacle, a small improvement in the safety distance can lead to a significant score gain, thus better guiding the model to avoid dangerous areas.

\textbf{Goal-Alignment Score ($S_{tar}$).} This score is used to evaluate whether the candidate points of orientation align with the user's final goal. Let the UAV's current position be $p_{curr}$, the candidate point be $p_c$, and the global final target point be $p_{tar}$.

We define a candidate vector, $\vec{v}_{\text{cand}} = p_c - p_{\text{curr}}$, which represents the direction of movement from the current position to the candidate point. Similarly, we define a target vector, $\vec{v}_{\text{tar}} = p_{\text{tar}} - p_{\text{curr}}$, which represents the direction from the current position to the final target.
We measure the directional consistency of these two vectors by calculating the cosine similarity between them. To normalize the score to the $[0, 1]$ interval, we perform a linear transformation:

\begin{equation}
    S_{tar} = \frac{1}{2} \left( \frac{\vec{v}_{cand} \cdot \vec{v}_{tar}}{\|\vec{v}_{cand}\|_2 \cdot \|\vec{v}_{tar}\|_2} + 1 \right)
    \label{eq:target_score}
\end{equation}

When the candidate direction is exactly the same as the target direction, $S_{\text{tar}}=1$; when they are orthogonal, $S_{\text{tar}}=0.5$; when the directions are completely opposite, $S_{\text{tar}}=0$.

\textbf{Forward-Progress Score ($S_{prog}$).}
This score is designed to encourage the UAV to make effective forward movements and avoid wandering in place or moving in a small range. We use the Euclidean distance $d_{prog}$ between the candidate point $p_c$ and the current position $p_curr$ to quantify the size of the forward progress:

\begin{equation}
d_{prog} = \| p_c - p_{curr} \|_2
\label{eq:progress_score1}
\end{equation}

However, directly using distance as a score has a drawback: a candidate point that is extremely far away but potentially dangerous will receive an excessively high score, thus affecting the balance of the decision. To solve this problem, we introduce the hyperbolic tangent function ($\tanh$) to normalize the distance:

\begin{equation}
    S_{prog} = \tanh\left(\frac{d_{prog}}{\alpha}\right)
    \label{eq:progress_score}
\end{equation}

Because the tanh function exhibits diminishing returns, it smoothly maps any distance value to the range $[0, 1)$. When the distance traveled is short, the score improves significantly; however, as the distance becomes longer, the score growth levels off. $\alpha$ is a scaling hyperparameter (set to 10.0 in our experiments) that controls how quickly the score reaches saturation. This design rewards effective progress while preventing the Forward-Progress Score from disproportionately dominating the final score.

\textbf{Final Adjudication.} After all candidate points are evaluated by the above scoring function, we will select the point with the highest total score as the final navigation decision. The best candidate point $w_{best}$ is selected by the following formula:

\begin{equation}
    w_{best} = \underset{c'_i \in C_{refined}}{\operatorname{arg\,max}} \left( Score_{total}({c'_i}) \right)
    \label{eq:final_selection}
\end{equation}
where $c'_i$ is the refined candidate point generated by the model after M rounds of self-correction. In UAV navigation tasks, this multi-heuristic optimization mechanism can help the UAV finally select a robust solution that achieves a good balance between safety, goal-alignment, and forward-progress.

\begin{table*}[t]
\begin{center}
\setlength{\tabcolsep}{1.5mm}
\begin{tabular}{@{}c@{}cccccccccccc@{}} 
\toprule
\multirow{2}{*}{\textbf{Method}} & \multicolumn{12}{c}{\textbf{Test Set}} \\ 
\cmidrule(l){2-13} 
 & \multicolumn{4}{c}{\textbf{TS}} & \multicolumn{4}{c}{\textbf{UO}} & \multicolumn{4}{c}{\textbf{UM}} \\
 & \textbf{NE ↓} & \textbf{SR ↑} & \textbf{OSR ↑} & \textbf{SPL ↑} & \textbf{NE ↓} & \textbf{SR ↑} & \textbf{OSR ↑} & \textbf{SPL ↑} & \textbf{NE ↓} & \textbf{SR ↑} & \textbf{OSR ↑} & \textbf{SPL ↑} \\ 
\midrule
Random & 222.20 & 0.14 & 0.21 & 0.07 & 260.14 & 0.16 & 0.16 & 0.16 & 202.98 & 0.00 & 0.00 & 0.00 \\
Fixed Action~\cite{garcia-aunon2019behavior} & 188.61 & 2.27 & 8.16 & 1.40 & 212.84 & 3.66 & 9.54 & 2.16 & 180.47 & 0.52 & 2.61 & 0.39 \\
CMA~\cite{yang2022vision} & 135.73 & 8.37 & 18.72 & 7.90 & 155.79 & 9.06 & 16.06 & 8.68 & 141.68 & 2.30 & 10.02 & 2.16 \\
TravelUAV~\cite{Wang2024} & 110.99 & 22.94 & 44.92 & 19.50 & 124.79 & 24.48 & 41.49 & 21.53 & 143.08 & 5.32 & 15.45 & 4.61 \\
Ours & \textbf{106.32} & \textbf{24.96} & \textbf{47.39} & \textbf{20.93} & \textbf{121.11} & \textbf{25.76} & \textbf{43.08} & \textbf{22.66} & \textbf{135.58} & \textbf{6.26} & \textbf{16.81} & \textbf{5.26} \\ 
\bottomrule
\end{tabular}
\end{center}
\caption{\textbf{Quantitative comparison with existing methods on different test sets.} We tested our approach on three different subsets: TS (Test Seen), UO (Unseen Object), and UM (Unseen Map), comparing it with four different baselines. Random refers to random path selection without any planning capabilities. Fixed Action uses heuristic methods to directly map language cues to pre-defined discrete maneuvers. CMA is a classic VLN model that uses cross-modal attention to align visual and language information for navigation. TravelUAV is a method based on a large language model that employs a hierarchical planning strategy, first generating macroscopic target points and then refining the specific flight trajectory. SR is the mission success rate, OSR is the probability that the UAV is able to maintain the correct overall heading but fails to reach the target, SPL is the flight path efficiency when the mission is successful, and NE is the distance difference between the UAV's final position and the target.
The results of Random, Fixed Action, and CMA are cited from Wang \emph{et al}.~\cite{Wang2024}, and the results of TravelUAV are our replication values of the methods in that paper.
}
\label{tab:table1}
\end{table*}

\begin{table}[t]
\begin{center}
\setlength{\tabcolsep}{1mm}
\begin{tabular}{@{}lcccccc@{}}
\toprule
\textbf{Method} & \textbf{Par} & \textbf{Ser} & \textbf{NE ↓} & \textbf{SR ↑} & \textbf{OSR ↑} & \textbf{SPL ↑} \\
\midrule
TravelUAV~\cite{Wang2024} & 1 & 1 & 110.99 & 22.94 & 44.92 & 19.50 \\
\hline
\multirow{3}{*}{Ours} & 2 & 1 & 111.29 & 23.06 & 44.71 & 19.83 \\
 & 3 & 1 & 108.16 & 23.27 & 45.13 & 20.37 \\
 & 3 & 2 & \textbf{106.32} & \textbf{24.96} & \textbf{47.39} & \textbf{20.93} \\
\bottomrule
\end{tabular}
\end{center}
\caption{\textbf{Parallel number comparison experiment.} Par represents the number of candidate points generated or evaluated simultaneously. Ser represents the number of times a candidate point is serially considered. The serial depth Ser is fixed at 1, and the parallelism Par is incremented from 1 to 3 to observe the impact of the parallelism on various metrics.  }
\label{tab:result1}
\end{table}
\subsection{Experimental replication and evaluation}
The experimental reproduction process in this article covers the complete implementation process, from environment configuration to performance evaluation. To avoid dependency conflicts, we used Conda to create a separate Python 3.10 virtual environment named llamauav, and installed the PyTorch 2.0.1 framework and its corresponding computer vision library within this environment. We also installed the project-specific LLaMA-UAV model dependencies, including core modules built from source code and optimization components to improve training efficiency, such as the Ninja compiler and a specific version of the Flash-Attention library. Furthermore, we installed the remaining necessary dependencies according to the requirements.txt file provided by the project and performed compatibility checks on the AirSim Python client to ensure proper interaction between the simulation module and the model training process.

For data preparation, we obtained the TRAVEL UAV dataset~\cite{Wang2024} and its official classification information from the Hugging Face platform. After obtaining the raw data, we need to standardize it using the project's preprocessing tools. This not only generates trajectory description files in a unified format, but also converts multi-view camera images into a tensor format recognizable by the model. This creates a structured and standardized data input system, laying the foundation for subsequent model training and evaluation.

The model preparation phase involves loading and organizing multiple pretrained weights. The core model uses the fine-tuned Vicuna-7B and layered trajectory decoder, the visual encoder uses the EVA-ViT-G and Q-Former modules, and the object detection component is based on the GroundingDINO model. These weight files must be placed in the model\_zoo folder according to the project's specified directory hierarchy to ensure correct model invocation and initialization.

The experimental simulation platform is built on AirSim, which provides a high-fidelity virtual environment package covering various terrain types, including typical scenes such as urban blocks, European-style buildings, and parks. Before evaluation, the simulation server program must be started and a stable communication channel with the simulation environment must be established by setting the service port and the environment's root directory path to ensure smooth execution of the UAV navigation mission. Performance evaluation is automatically completed through standardized scripts. The evaluation system supports DAgger training mode and closed-loop evaluation mode. After running, it can automatically execute the entire process of navigation trajectory generation, simulation data collection, and performance indicator calculation, thereby achieving quantitative verification of the model's comprehensive performance under multiple scenarios.



\begin{table}[t]
\begin{center}
\setlength{\tabcolsep}{1mm}
\begin{tabular}{@{}lcccccc@{}}
\toprule
\textbf{Method} & \textbf{Par} & \textbf{Ser} & \textbf{NE ↓} & \textbf{SR ↑} & \textbf{OSR ↑} & \textbf{SPL ↑} \\
\midrule
TravelUAV~\cite{Wang2024} & 1 & 1 & 110.99 & 22.94 & 44.92 & 19.50 \\
\hline
\multirow{2}{*}{Ours} & 1 & 2 & 107.23 & 24.54 & 46.40 & 20.89 \\
 & 3 & 2 & \textbf{106.32} & \textbf{24.96} & \textbf{47.39} & \textbf{20.93} \\
\bottomrule
\end{tabular}
\end{center}
\caption{\textbf{Serial depth comparison experiment.} The parallel number Par is fixed at 1, and the serial depth Ser is increased from 1 to 2 to observe the impact of the serial depth on various indicators.}
\label{tab:result2}
\end{table}

\section{Experiment}

\subsection{Experimental Setup and Results}


\begin{figure*}[t] 
    \centering
    \includegraphics[width=0.9\textwidth]{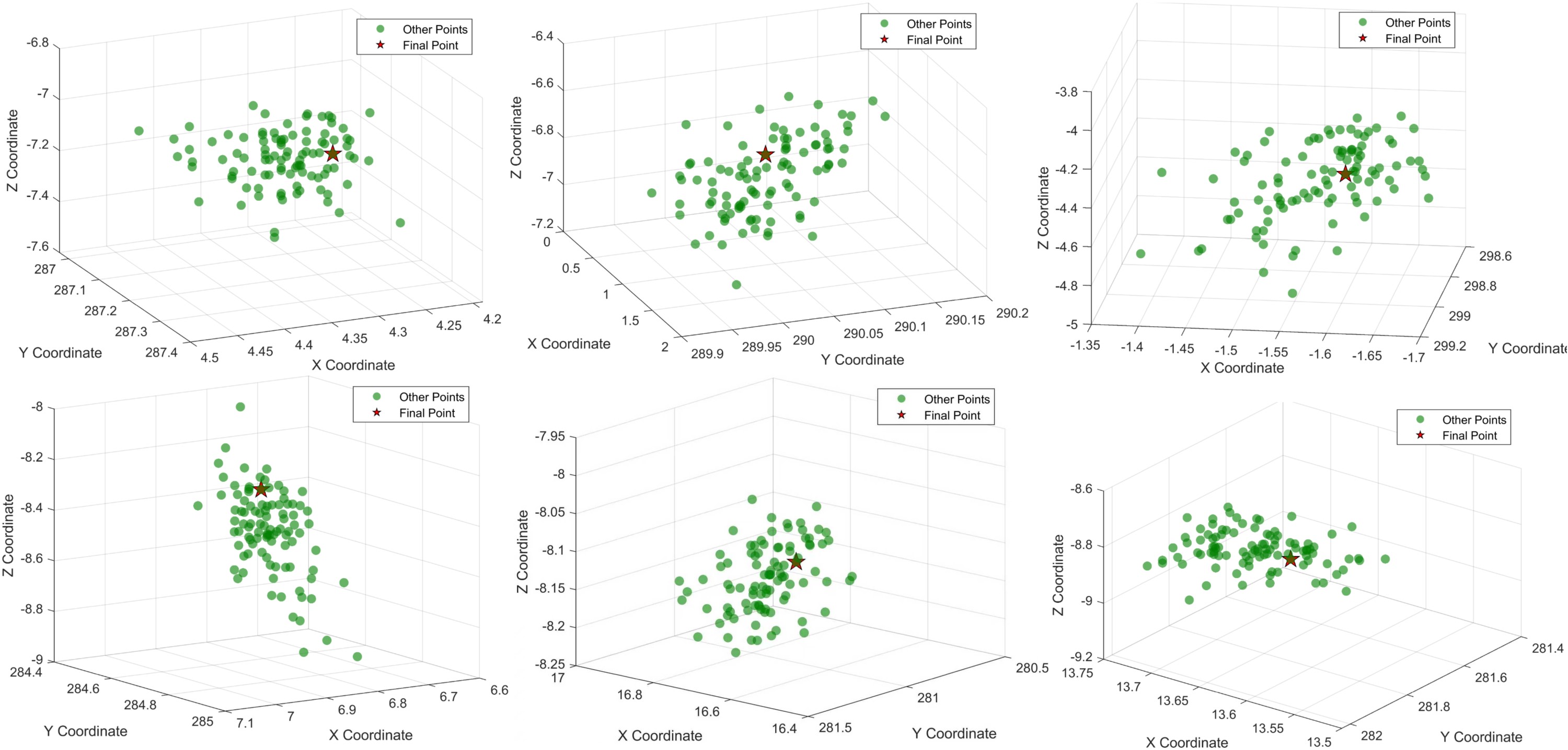} 
    \caption{
    \textbf{A scatter plot of the model's first phase of self-reflection after 100 parallel reflections.} The green dots represent 100 possible solutions, and the red stars represent the next pathpoints selected by a multi-criteria scoring function.
    }
    \label{fig:sandian}
\end{figure*}

We quantitatively evaluate our approach on three standard test sets: 1) TS (Test Seen) uses objects and scenes seen during training. 2) UO (Unseen Object) contains new objects not seen during training. 3) UM (Unseen Map) is tested on new scenes not seen during training. We compare against several baseline models, including: 1) Random: The UAV chooses a completely random route pose at each decision point, without any planning or understanding. 2) Fixed Action: This is a simple, rule-based, non-learning policy. It maps keywords in commands to a set of predefined, discrete actions, such as ``cruise'' meaning to move forward 5 meters. 3) CMA: This is a classic visual-language navigation model based on a cross-modal attention mechanism. It learns to align the semantics of language commands with the features of visual observations and make decisions. 4) TravelUAV: This is a multimodal navigation method based on a large language model that can simultaneously process image and target commands to tackle complex navigation tasks. Its core is the hierarchical trajectory generation mechanism, that is, the long-distance macro target position is first planned by a large multimodal model, and then the path decoder is combined with the front view image of the UAV to generate a detailed flight trajectory sequence.

The model offers three optional assistant modes: L1, L2, and L3, which provide different navigation assistance to the UAV, with assistance levels decreasing in intensity. Specifically, the L1 assistant provides continuous, high-frequency guidance to ensure the UAV stays on course, the L2 assistant provides low-frequency corrections only when the UAV deviates from its path or encounters difficulties, and the L3 assistant provides minimal obstacle avoidance instructions only when a collision is imminent. In the following experiments, we uniformly select the L1 mode to verify the effectiveness of our framework.

\begin{figure}[t] 
    \centering
    \includegraphics[width=0.5\textwidth]{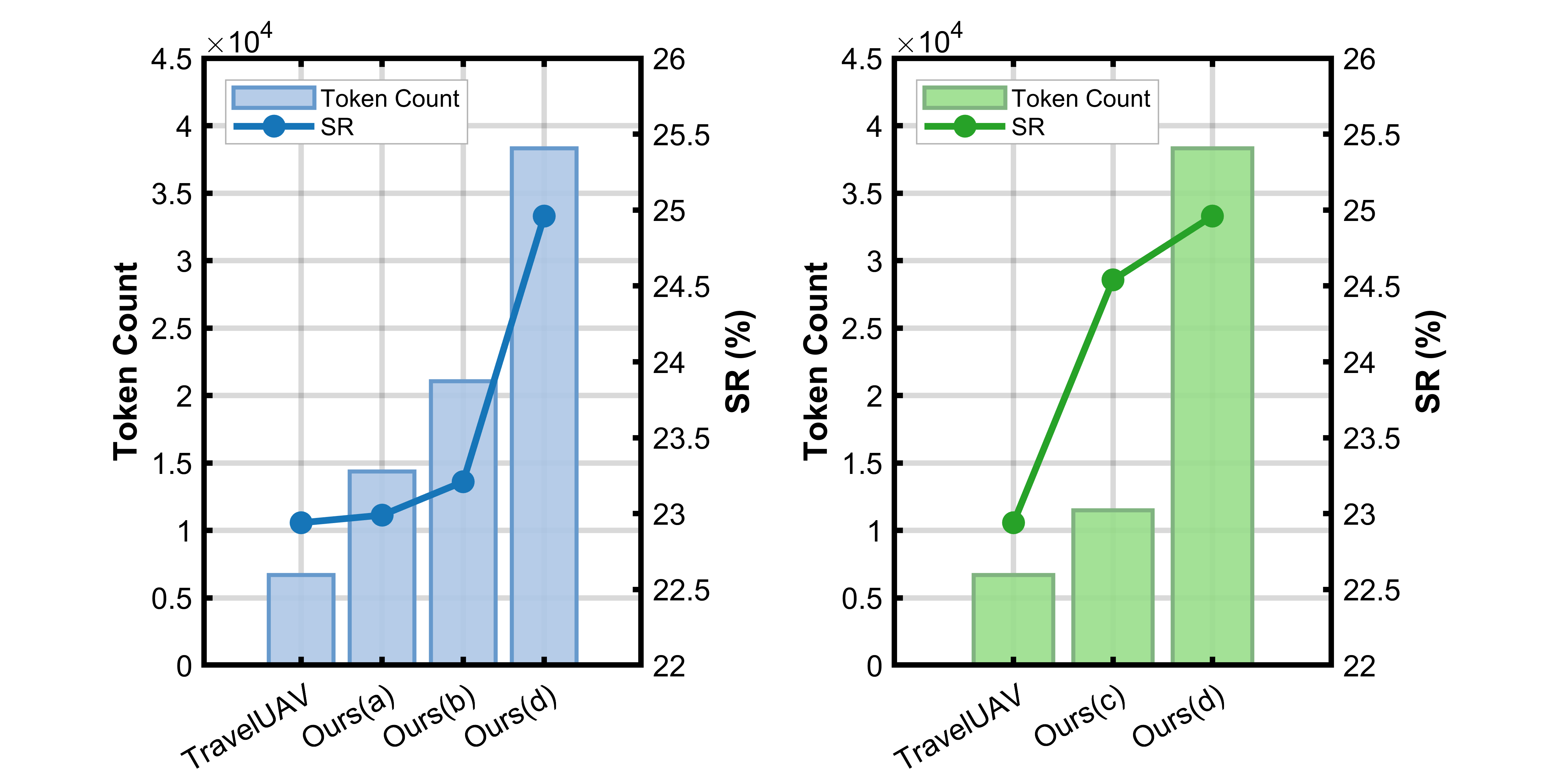} 
    \caption{
\textbf{Correlation between the number of inference tokens and model accuracy (measured in SR).} This figure shows the relationship between the number of tokens generated by the model and SR under different test conditions. The figure uses a bar chart to represent the number of tokens generated during the generation phase (left vertical axis), and a broken line to represent the corresponding SR (right vertical axis), visually depicting the correlation between computational overhead and model performance. Each data point represents a different configuration: TravelUAV has a parallelism of 1 and a serial depth of 1; Ours (a) has a parallelism of 2 and a serial depth of 1; Ours (b) has a parallelism of 3 and a serial depth of 1; Ours (c) has a parallelism of 1 and a serial depth of 2; Ours (d) has a parallelism of 3 and a serial depth of 2. It can be observed that as the number of tokens generated during the inference phase increases, the model's SR continues to improve, indicating that a higher computational budget can significantly improve the model's inference accuracy.
    }
    \label{fig:figure}
\end{figure}


\begin{figure*}[t]
    \centering
    \includegraphics[width=0.9\textwidth]{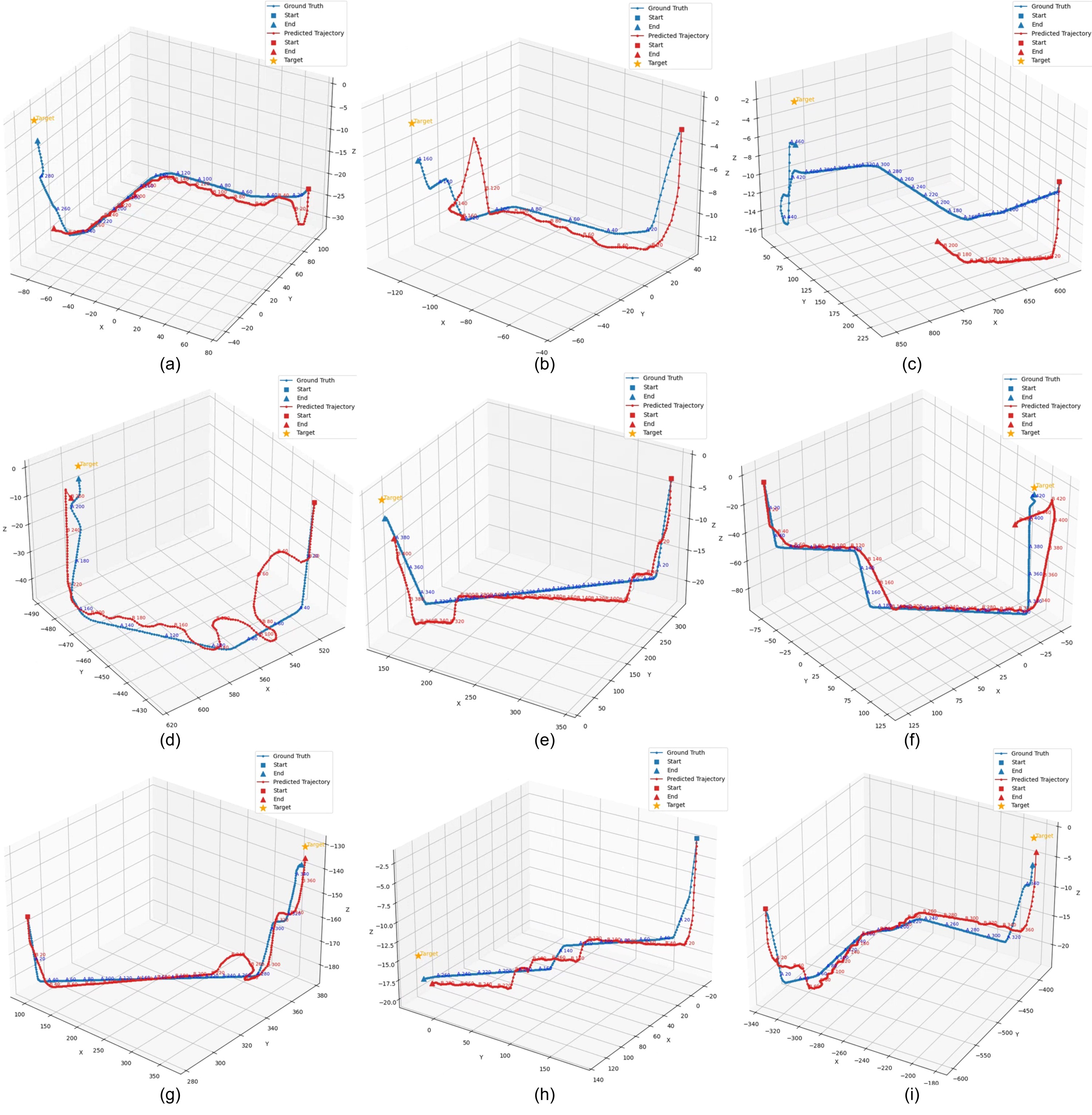}
    \caption{%
        \textbf{Visualization of trajectories under different outcome categories.} This figure visualizes the comparison between the 3D flight paths generated by model evaluation and the actual trajectories. Blue lines represent the ground truth, while red lines indicate the trajectories predicted and executed by the VLMs. The starting point (Start) is marked by a square, and the endpoint (End) is marked by a triangle. The final target point is marked by a yellow star.
    }
    \label{fig:visual}
\end{figure*}

As shown in Table~\ref{tab:table1}, our proposed method achieves state-of-the-art performance on all standard test sets, primarily attributable to its three-stage \emph{Explore--Refine--Select} reasoning framework. This framework first broadens the search scope through parallel exploration to prevent the model from prematurely falling into a single suboptimal solution. Subsequently, each preliminary solution is critically reflected upon and refined through serial optimization. Finally, a multi-criteria scoring function quantitatively evaluates all optimized solutions based on safety, goal consistency, and task progress to ensure the selection of the overall optimal solution. The synergistic effect of these three stages ultimately translates into higher SR, OSR, SPL, and lower NE metrics in navigation tasks. Specifically, on the TS test set, our method improves SR by 2.02\%, SPL and OSR by 1.43\% and 2.47\%, respectively, and reduces NE by 4.67\%. On the UO and UM test sets, SR is improved by 1.28\% and 0.94\%, proving that the framework has a certain generalization ability to new objects and environments.

We observe that the gains on UO and UM are smaller than those on TS. TS contains 1410 trajectories from objects and scenes that are already present in training. In contrast, UO contains 629 trajectories and includes 13 objects that do not appear in training. UM contains 958 trajectories and includes 2 scenes that do not appear in training. These changes introduce distribution shifts, so it is reasonable to expect that the task becomes harder than TS. For UO, unseen objects change visual cues in the scene, which affect perception and visual grounding. For UM, unseen scenes change the spatial layout and free-space patterns, which make waypoint selection less stable. 



\subsection{Analysis of Parallel and Serial Scaling Dimensions}

As shown in Table~\ref{tab:result1}, experimental results quantify the positive correlation between model performance and the number of parallel candidates. We fixed the serial depth (Ser) at 1 and gradually increased the number of parallel candidates (Par). The data shows that increasing Par from 1 (baseline) to 3 leads to modest improvements in SR, SPL, OSR, and NE.

This trend can be attributed to the inherent randomness and uncertainty in the generation of large language models. For any given complex instruction, VLMs can generate a large number of seemingly plausible but highly detailed outputs. A single generation (Par=1) can easily accidentally sample a suboptimal initial candidate from this rich distribution of possibilities. Increasing the number of parallelism (Par=2 and Par=3) allows the model to sample a wider range of possibilities. Due to the uncertainty of model generation, a single attempt may result in a suboptimal solution. By broadening the search range, we are able to explore more possibilities, thereby increasing the probability of capturing the global optimal candidate.

As shown in Table~\ref{tab:result2}, the experimental results quantify the positive correlation between model performance and serial depth.When we fix the number of Par at 1, increasing the Ser from 1 to 2 leads to overall performance improvements in all reported metrics. This result is a direct manifestation of ``test time scaling'', which involves allocating a larger computational budget at inference time in exchange for improved performance. The initial candidate points generated when Ser=1 can be considered the model’s initial plan. By introducing a second revision step (Ser=2), we force the model to perform a critical self-correction. This additional step consumes more computational resources, effectively increasing the model’s thinking time. This allows the model to better identify and correct potential flaws in its initial plan, such as suboptimal paths or overlooked safety hazards.

A comprehensive analysis of the experimental results shows that combining parallel exploration with serial revision maximizes navigation performance. As shown in Table~\ref{tab:result1} and Table~\ref{tab:result2}, our configuration (Par=3, Ser=2) achieves the best results across all evaluation metrics, outperforming any method that relies solely on a single expansion dimension. This reveals a significant synergistic effect: parallel exploration first ensures the model considers a diverse range of initial policies, while the subsequent serial revision deeply optimizes these policies. It is this combination of breadth and depth that ultimately leads to the most reliable and efficient navigation decisions.

\subsection{Visualization of VLMs' Planning Uncertainty}

To visually demonstrate the VLMs' uncertainty during planning, we visualize the distribution of candidate waypoints generated by the parallel exploration phase based on the TS dataset in Figure~\ref{fig:sandian}. Given the same initial state, the candidate points generated by the model are scattered in clusters across three-dimensional space, rather than converging at a single point. It can be seen that when VLMs makes decisions in the face of complex instructions, it will generate the probabilistic nature of multiple candidate points.
Traditional single-step reasoning methods randomly select one of these candidate points, leading to unstable decisions. Our parallel exploration strategy is designed to systematically address this uncertainty: by capturing this diverse set of possibilities, it provides a comprehensive candidate set for the subsequent scoring and prioritization phase, thereby selecting a more reasonable and safe final decision.

\subsection{Computational Budget vs. Navigation Accuracy}

Experimental data confirms a fundamental premise of the test-time scaling framework: allocating more computational budget for inference (measured by token consumption) is often associated with better navigation decision quality. Experimental results based on the TS dataset confirm this positive correlation. As shown on the left side of Figure~\ref{fig:figure}, as the number of tokens consumed increases from 6,705 to 38,315, the SR steadily improves from 22.94\% to 24.96\%. The data on the right side of the figure further confirms this positive correlation. This result demonstrates that allocating more computational resources to model inference translates into more reliable navigation performance.
\subsection{Qualitative Analysis}
To complement the quantitative findings, we performed a qualitative analysis of the UAV's navigation behavior. Figure~\ref{fig:visual} visualizes several trajectory cases generated by the model and compares them with the ground truth. These cases are categorized into three different result types: failure, oracle success, and success. (a-c) show failure cases where the UAV did not reach the target point due to reasons such as collision. (d-f) show oracle success cases where the UAV was able to maintain the correct general heading, but ultimately landed near the target area but did not successfully reach the target. (g-i) show success cases where the UAV successfully reached the target.

\section{Conclusion}
In this paper, we propose a simple yet effective test-time scaling framework to address the single-step decision-making limitations of current visual language model-based UAV navigation methods. This framework employs a three-stage \emph{Explore--Refine--Select} pipeline: First, a diverse set of candidate waypoints is generated. Each candidate is then optimized through a self-correction mechanism. Finally, the optimal path is selected using a multi-criteria scoring function, significantly enhancing the reasoning capabilities of the frozen VLMs. Experimental results demonstrate that our approach outperforms state-of-the-art TravelUAV baselines on key metrics across both seen and unseen object and map test sets. Furthermore, we demonstrate that combining parallel and serial scaling yields superior synergy and confirm a positive correlation between computational budget and navigation accuracy.
The key contribution of this research is that it demonstrates that increasing the number of deliberation steps in the reasoning process can effectively improve the reliability of UAV navigation, which has significant safety implications for real-world applications. In the future, we plan to explore how the model can adaptively adjust the breadth and depth of deliberation based on task difficulty and aim to deploy this approach in real-world UAV systems for validation and application.

\section{Acknowledgement}
This work was supported in part by the Key Research and Development Program of Shandong Province under Grant No.~2025CXGC010901.

{\small
\bibliographystyle{cvm}
\bibliography{cvmbib}

@article{bai2025qwen2,
  title={Qwen2. 5-vl technical report},
  author={Bai, Shuai and \etal},
  journal={arXiv preprint arXiv:2502.13923},
  year={2025}
}

@inproceedings{bednavr2022deployment,
  title={Deployment of reliable visual inertial odometry approaches for unmanned aerial vehicles in real-world environment},
  author={Bedn{\'a}{\v{r}}, Jan and \etal},
  booktitle={2022 International Conference on Unmanned Aircraft Systems (ICUAS)},
  pages={167--176},
  year={2022},
  organization={IEEE}
}

@article{cai2025flightgpt,
  title={FlightGPT: Towards Generalizable and Interpretable UAV Vision-and-Language Navigation with Vision-Language Models},
  author={Cai, Hengxing and \etal},
  journal={arXiv preprint arXiv:2505.12835},
  year={2025}
}

@article{chang2023review,
  title={A review of UAV autonomous navigation in GPS-denied environments},
  author={Chang, Yingxiu and \etal},
  journal={Robotics and Autonomous Systems},
  volume={170},
  pages={104533},
  year={2023},
  publisher={Elsevier}
}

@article{chen2024far,
  title={How far are we to gpt-4v? closing the gap to commercial multimodal models with open-source suites},
  author={Chen, Zhe and \etal},
  journal={Science China Information Sciences},
  volume={67},
  number={12},
  pages={220101},
  year={2024},
  publisher={Springer}
}

@article{diez2024time,
  title={Time-based UWB localization architectures analysis for UAVs positioning in industry},
  author={D{\'\i}ez-Gonz{\'a}lez, Javier and \etal},
  journal={Ad Hoc Networks},
  volume={157},
  pages={103419},
  year={2024},
  publisher={Elsevier}
}

@article{Fan2022,
  author    = {Fan, Y. and \etal},
  title     = {Aerial Vision-and-Dialog Navigation},
  journal   = {arXiv preprint arXiv:2205.12219},
  year      = {2022}
}

@article{garcia-aunon2019behavior,
  title={Behavior-based control for an aerial robotic swarm in surveillance missions},
  author={Garcia-Aunon, P. and \etal},
  journal={Sensors},
  volume={19},
  number={20},
  pages={4584},
  year={2019}
}

@article{guo2025autonomous,
  title={Autonomous UAV last-mile delivery in urban environments: A survey on deep learning and reinforcement learning solutions},
  author={Guo, Jingrui and \etal},
  journal={Control Engineering Practice},
  volume={165},
  pages={106491},
  year={2025},
  publisher={Elsevier}
}

@article{hanover2024autonomous,
  title={Autonomous drone racing: A survey},
  author={Hanover, Drew and \etal},
  journal={IEEE Transactions on Robotics},
  volume={40},
  pages={3044--3067},
  year={2024},
  publisher={IEEE}
}

@article{jia2024drone,
  title={Drone-NeRF: Efficient NeRF based 3D scene reconstruction for large-scale drone survey},
  author={Jia, Zhihao and \etal},
  journal={Image and Vision Computing},
  volume={143},
  pages={104920},
  year={2024},
  publisher={Elsevier}
}

@article{kaufmann2023champion,
  title={Champion-level drone racing using deep reinforcement learning},
  author={Kaufmann, Elia and \etal},
  journal={Nature},
  volume={620},
  number={7976},
  pages={982--987},
  year={2023},
  publisher={Nature Publishing Group UK London}
}

@article{li2024llava,
  title={Llava-onevision: Easy visual task transfer},
  author={Li, Bo and \etal},
  journal={arXiv preprint arXiv:2408.03326},
  year={2024}
}

@article{li2024llava0,
  title={Llava-next-interleave: Tackling multi-image, video, and 3d in large multimodal models},
  author={Li, Feng and \etal},
  journal={arXiv preprint arXiv:2407.07895},
  year={2024}
}

@article{li2025tokenpacker,
  title={Tokenpacker: Efficient visual projector for multimodal llm},
  author={Li, Wentong and \etal},
  journal={International Journal of Computer Vision},
  pages={1--19},
  year={2025},
  publisher={Springer}
}

@article{liu2024cooperative,
  title={Cooperative relative localization in mav swarms with ultra-wideband ranging},
  author={Liu, Changrui and \etal},
  journal={arXiv preprint arXiv:2405.18234},
  year={2024}
}

@article{liu2023visual,
  title={Visual instruction tuning},
  author={Liu, Haotian and \etal},
  journal={Advances in neural information processing systems},
  volume={36},
  pages={34892--34916},
  year={2023}
}

@inproceedings{liu2023aerialvln,
  title={Aerialvln: Vision-and-language navigation for uavs},
  author={Liu, Shubo and \etal},
  booktitle={Proceedings of the IEEE/CVF International Conference on Computer Vision},
  pages={15384--15394},
  year={2023}
}

@inproceedings{Madaan2023,
  author    = {Madaan, A. and \etal},
  title     = {Self-Refine: Iterative Refinement with Self-Feedback},
  booktitle = {Advances in Neural Information Processing Systems},
  volume    = {36},
  year      = {2023}
}

@article{montello2025survey,
  title={A Survey on Dynamic Neural Networks: from Computer Vision to Multi-modal Sensor Fusion},
  author={Montello, Fabio and \etal},
  journal={arXiv preprint arXiv:2501.07451},
  year={2025}
}

@article{muennighoff2025s1,
  title={s1: Simple test-time scaling},
  author={Muennighoff, Niklas and \etal},
  journal={arXiv preprint arXiv:2501.19393},
  year={2025}
}

@article{oyinlola2025reinforcement,
  title={Reinforcement Learning for Autonomous Point-to-Point UAV Navigation},
  author={Oyinlola, Salim and \etal},
  journal={arXiv preprint arXiv:2509.13943},
  year={2025}
}

@inproceedings{pritzl2024drones,
  title={Drones guiding drones: Cooperative navigation of a less-equipped micro aerial vehicle in cluttered environments},
  author={Pritzl, V{\'a}clav and \etal},
  booktitle={2024 IEEE/RSJ International Conference on Intelligent Robots and Systems (IROS)},
  pages={10597--10604},
  year={2024},
  organization={IEEE}
}

@article{pritzl2023fusion,
  title={Fusion of visual-inertial odometry with LiDAR relative localization for cooperative guidance of a micro-scale aerial vehicle},
  author={Pritzl, V{\'a}clav and \etal},
  journal={arXiv preprint arXiv:2306.17544},
  year={2023}
}

@article{snell2024scaling,
  title={Scaling llm test-time compute optimally can be more effective than scaling model parameters, 2024},
  author={Snell, Charlie and \etal},
  journal={URL https://arxiv. org/abs/2408.03314},
  volume={20},
  year={2024}
}

@article{song2023efficient,
  title={Efficient evaluation methods for neural architecture search: A survey},
  author={Song, Xiaotian and \etal},
  journal={arXiv preprint arXiv:2301.05919},
  year={2023}
}

@inproceedings{song2025towards,
  title={Towards long-horizon vision-language navigation: Platform, benchmark and method},
  author={Song, Xinshuai and \etal},
  booktitle={Proceedings of the Computer Vision and Pattern Recognition Conference},
  pages={12078--12088},
  year={2025}
}

@inproceedings{su2025learning,
  title={Learning Fine-Grained Alignment for Aerial Vision-Dialog Navigation},
  author={Su, Yifei and \etal},
  booktitle={Proceedings of the AAAI Conference on Artificial Intelligence},
  volume={39},
  pages={7060--7068},
  year={2025}
}

@article{wang2024qwen2,
  title={Qwen2-vl: Enhancing vision-language model's perception of the world at any resolution},
  author={Wang, Peng and \etal},
  journal={arXiv preprint arXiv:2409.12191},
  year={2024}
}

@article{Wang2024,
  author    = {Wang, X. and \etal},
  title     = {Towards Realistic UAV Vision-Language Navigation: Platform, Benchmark, and Methodology},
  journal   = {arXiv preprint arXiv:2410.07087},
  year      = {2024}
}

@inproceedings{Wei2022,
  author    = {Wei, J. and \etal},
  title     = {Chain-of-thought prompting elicits reasoning in large language models},
  booktitle = {Advances in Neural Information Processing Systems},
  volume    = {35},
  pages     = {24824--24837},
  year      = {2022}
}

@article{wu2025aeroduo,
  title={AeroDuo: Aerial Duo for UAV-based Vision and Language Navigation},
  author={Wu, Ruipu and \etal},
  journal={arXiv preprint arXiv:2508.15232},
  year={2025}
}

@article{xiao2024beyond,
  title={Beyond model adaptation at test time: A survey},
  author={Xiao, Zehao and Snoek, Cees GM},
  journal={arXiv preprint arXiv:2411.03687},
  year={2024}
}

@article{yang2022vision,
  title={Vision-language pre-training with triple contrastive learning},
  author={Yang, J. and \etal},
  journal={Proceedings of the IEEE/CVF Conference on Computer Vision and Pattern Recognition (CVPR)},
  pages={15671--15680},
  year={2022}
}

@article{Yao2023,
  author    = {Yao, S. and \etal},
  title     = {Tree of thoughts: Deliberate problem solving with large language models},
  journal   = {arXiv preprint arXiv:2305.10601},
  year      = {2023}
}

@inproceedings{ye2025atp,
  title={Atp-llava: Adaptive token pruning for large vision language models},
  author={Ye, Xubing and \etal},
  booktitle={Proceedings of the Computer Vision and Pattern Recognition Conference},
  pages={24972--24982},
  year={2025}
}

@article{zhang2025grounded,
  title={Grounded Vision-Language Navigation for UAVs with Open-Vocabulary Goal Understanding},
  author={Zhang, Yuhang and \etal},
  journal={arXiv preprint arXiv:2506.10756},
  year={2025}
}

@inproceedings{zhong2024safer,
  title={A safer vision-based autonomous planning system for quadrotor uavs with dynamic obstacle trajectory prediction and its application with llms},
  author={Zhong, Jiageng and \etal},
  booktitle={Proceedings of the IEEE/CVF winter conference on applications of computer vision},
  pages={920--929},
  year={2024}
}
}

\end{document}